# NoFake at CheckThat! 2021: Fake News Detection Using BERT

Sushma Kumari

*Independent Researcher*

**Abstract**
Much research has been done for debunking and analysing fake news. Many researchers study fake news detection in the last year, but many are limited to social media data. Currently, multiples fact-checkers are publishing their results in various formats. Also, multiple fact-checkers use different labels for the fake news, making it difficult to make a generalisable classifier. With the merge classes, the performance of the machine model can be enhanced. This domain categorisation will help group the article, which will help save the manual effort in assigning the claim verification. In this paper, we have presented BERT based classification model to predict the domain and classification. We have also used additional data from fact-checked articles. We have achieved a macro F1 score of 83.76 % for Task 3Aand 85.55 % for Task 3B using the additional training data.

## 1. Introduction

Data journalism is a new journalistic discipline that focuses mainly on data-driven research and presentation formats. However, a fundamental problem of data journalism and classical journalism is that much data of journalistic interest is only available in the unstructured form: as texts, tables and graphics in documents of various types (Word, PDF, e-mail, etc.) or on websites. Also, there is a lack of a centralised data hub for gathering information from different sources.

There is an increasing amount of fake news in the media, social media, and other web sources. Much research has been done for fake news detection and debunking of fake news [1]. In the last two decades, there is a tremendous increase in the spread of misinformation, which is also reflected by the number of fact-checking websites [1]. Fact-checking websites can help to investigate claims and assist citizens in determining whether the information used in an article is true or not. More than 213 fact-checking websites are working in 40+ languages across 100+ countries [2, 3, 4].

There is no standard protocol for fact-checking services across different fact-checkers, and they do not publish their proofed articles in a standard format, which leads to several conflicts. Shahi et al. [3] discuss the need to detect news articles potentially containing fake information.

In this paper, we discussed the method used for the fake news detection for the shared task at Checkthat!. The remainder of the paper is organised as related work describes the past work, task description gives an overview of the task, Experiment section emphasises on the experiment details and Conclusion, and Future Work focuses on the conclusion and future aspect of the work.

## 2. Related Work

Fact-checking is a damage control method that is both essential and not scalable. It might be hard to take the human component out of the picture any time soon. But still, automatic fake news classification could help reduce the workload for the fact-checkers. The fake content is spread in multiple formats; many of them are repurposed by changing the text, location, etc. Fake news detection is a complex problem. Research has been done on fake news detection using social media data like tweets [2], YouTube videos [3], but less research has focused on the news articles. One of the primary reasons is the lack of corpus o news articles. Fake news is spread in several domains like crime, election, economy.

There is a lack of corpus to train Machine Learning based on fact-checking. In the last few years, a collection of small datasets related to fact-checking has been published. These datasets are a mixture of fake news topics, For instance, US election 2016[5]. Still, fact-checking is dominant in the English language and few sources like Snopes, Politifact etc.

Multi-FC corpus describes the different kinds of fact-checking datasets available and their limitations[6]. The authors have come up with a multi-domain, evidence-based fact-checking dataset. They have also described the metadata of Fact-check articles. [7] describes the task of fact-checking and construction of dataset using the process used by journalists. Several methods are published for automatic detection of fake news, Pérez-Rosas et al. discuss the automatic detection of fake news and linguists differences in false and legitimate content. [9] analyse, compare and summarise the several different methods available for fake news detection. The authors give an overview of methods that have been tested for fake news detection. Research has been done to look inside on the feature of a news story in the modern diaspora along with different kinds of story and their impact on people[10].

Both COVID-19 pandemic and infodemic spread in parallel, Mesquita et al.; proposes a framework to fight against fake medical news because fake news can intensify the effect of the COVID-19 pandemic [11]. All health workers, scientists, the government are trying to fight against fake medical news. In March 2020, several cases were discovered where people are consuming partially true information about COVID-19 on Facebook and WhatsApp without using the proper medical terms, which creates a panic about medicine and prevention for COVID-19 [12]. By analysing the search behaviour of people in Italy from January to March 2020 and found that a "large number of infodemic monikers were observed across Italy" [13].

During the time of the pandemic, fake news is spread all over the world in different languages. The fact-news is covered in different domains like origin and spread, conspiracy theory etc. There is a lack of resources that are multilingual and cross-domain and have been collected from multiple sources.

## 3. Task Description

The task was organised as CLEF 2021 - CheckThat! Lab Detecting Check-Worthy Claims, Previously Fact-Checked Claims, and Fake News [14, 15]. In task 3 of CheckThat!, the core idea of task is to classify fake news [16]. The task is defined as follows.

Task Definition: The first task is to classify the text of the news article into the four possible classes into true, partially true, false, or other (e.g., claims in dispute) and predict the topical domain of the article. The task is divided into two sub-task, they are: Subtask 3A: Multi-class fake news detection of news articles (English): Sub-task would be the classification of news articles in four classes as defined below:

- False - The main claim made in an article is untrue.
- Partially False - The main claim of an article is a mixture of true and false information. The article contains partially true and partially false information but cannot be considered as 100% true. It includes all articles in categories like partially false, partially true, mostly true, miscaptioned, misleading etc., as defined by different fact-checking services.
- True - This rating indicates that the primary elements of the main claim are demonstrably true.
- Other- An article cannot be categorised as true, false, or partially false due to lack of evidence about its claims. This category includes articles in dispute and unproven articles.

Subtask 3B: We often observed that the incoming flow of requests is from different topics such as cancer, COVID-19, diet, etc. With this categorisation, we can assign the respective person/expert to verify the content. It will help to save the manual effort in assigning the claim verification. The task was assigned for the following domain:

- Health: Article talking about health topics
- Election: The main focus of the article is election focused.
- Crime: These articles are mainly focused on crime.
- Climate: Articles which discuss climate issue
- Economy: Articles which discussed economic activity.
- Education: Articles focused on educational content.

## 4. Experiment

Transfer learning is a technique where a deep learning model trained on a large dataset performs similar tasks on another dataset. We call such a deep learning model a pre-trained model. The most renowned examples of pre-trained models are the computer vision deep learning models trained on the ImageNet dataset. So, it is better to use a pre-trained model as a starting point to solve a problem rather than building a model from scratch.

For training the model, We have to download the external data from the fact-checked articles. We have used the AMUSED framework [17], and FakeCovid [3] concept to merge the articles into four categories. We have downloaded 206,432 facts checked articles from more than 100 fact-checking websites. To gather the external dataset, we developed a Python-based crawler using Beautiful soup [18] to fetch all the news articles from the 92 different fact-checking websites; they are Poynter, Snopes, PolitiFact etc. Our crawler collects important information like the title of the news articles, name of the fact-checking websites, date of publication, the text of the news articles, and a class of news articles. For training, we have used the combination of text and text with the given dataset of the task. We have applied the basic cleaning process of natural language processing like removing special characters, punctuation marks etc.

To classify fake news articles, we used the state-of-the-art neural network language model BERT, which has been pre-trained on a large corpus to solve language processing tasks [19]. An essential advantage of BERT is that it can be fine-tuned for task-specific datasets and allows high text classification accuracy even for smaller datasets. In the context of fake news classification, BERT has already been applied for multiclass classification tasks, for example, on the Chinese social media platform Weibo, where it achieved considerable accuracy[20]. The randomisation of the data prevents seasonal patterns from being learned by the model. The randomisation of the data prevents seasonal patterns from being learned by the model. For BERT fine-tuning, the models for the videos were trained on for four epochs, and the models for the comments were trained on for three epochs with a learning rate of 2e-5. We set hidden units as 300 and training epoch as 200. Each training process continues until the restriction or validation loss is continued. The batch size is set to 10, and the learning rate is 0.001. After the individual prediction on the two test datasets, we evaluated the accuracy of the two models using the weighted F1 score. For classification, the model gave a macro F1 score of 83.76 % for Task 3A and 85.55 % for Task 3B.

## 5. Conclusion & Future Work

In this paper, We have presented a classification model for detecting the classification of fake news and its domain. Using the BERT model, the model performed well compared to the traditional machine learning technique. With the proposed model, the problem of fake news classification, domain identification will be addressed. As future work, the concept can be used for determining fake news on a large scale. Even the type of fake news can be embedded in the HTML content of the article [21].

Getting a fake news corpus is a challenging task, and with the limited dataset, it is hard to enhance the machine learning model's performance. With the external dataset, the performance of the classifier is increased. We also observed that different fact-checking websites are checking several duplicates of claims. Several old claims are repurposed as fake news. So, in future work, detecting similar claims would help find fake news classes using text matching.